# One-Shot GAN Generated Fake Face Detection

Hadi Mansourifar
University of Houston
Houston, USA
hmansourifar@uh.edu

Weidong Shi
University of Houston
Houston, USA
wshi3@central.uh.edu

*Abstract*—Fake face detection is a significant challenge for intelligent systems as generative models become more powerful every single day. As the quality of fake faces increases, the trained models become more and more inefficient to detect the novel fake faces, since the corresponding training data is considered outdated. In this case, robust One-Shot learning methods is more compatible with the requirements of changeable training data. In this paper, we propose a universal One-Shot GAN generated fake face detection method which can be used in significantly different areas of anomaly detection. The proposed method is based on extracting out-of-context objects from faces via scene understanding models. To do so, we use state of the art scene understanding and object detection methods as a pre-processing tool to detect the weird objects in the face. Second, we create a bag of words given all the detected out-of-context objects per all training data. This way, we transform each image into a sparse vector where each feature represents the confidence score related to each detected object in the image. Our experiments show that, we can discriminate fake faces from real ones in terms of out-of-context features. It means that, different sets of objects are detected in fake faces comparing to real ones when we analyze them with scene understanding and object detection models. We prove that, the proposed method can outperform previous methods based on our experiments on Style-GAN generated fake faces.

## I. INTRODUCTION

The invention of Generative Adversarial Networks (GANs) [20,17] has made it easy to synthesize realistic fake faces. The adversarial nature of GANs, lets the attackers to upgrade the discriminator loss function against fake image detection techniques. As the quality of fake faces increases, the trained models become more and more inefficient to detect the novel fake faces, since the corresponding training data is considered outdated. One viable solution to tackle this challenge is to decrease the dependency to training data to detect the fake faces. Traditional artificial intelligence works fine in case of huge training data. However, in case of scarce training data it fails to extract and generalize the knowledge [23]. Few-Shot Learning (FSL) [9,10,11,12] was proposed to meet the challenges of limited data. FSL can help the fake face detection systems to be more compatible against ever upgrading generative models. To the best of our knowledge, no effort has made toward One-Shot fake face detection and we take the first step in this direction. In this paper, we propose a universal FSL method based on out-of-context feature extraction which can be used in various applications from character recognition to fake face detection. First, we use Amazon Rekognition Console as state of the art scene understanding and object detection method to pre-process the input faces. The output of the model is a set of objects which are detected in the face. Second, we create a bag of words given all the detected objects per all data. This way, we transform each face image from numerical space into categorical domain. Thus, each image is represented by a sparse vector where each feature represents the confidence score related to each detected object in the face image. Our experiments show that, we can discriminate fake faces from real ones in terms of out-of-context features. It means that, different sets of objects are detected in fake faces comparing to real ones when we analyze them with scene understanding and object detection models. One important characteristic of proposed fake face detection is that, the test and training data are collected from completely different resources. In our research, the training instance is selected among synthesized instances by PG-GAN and test data are collected from *thispersondoesnotexist.com*. Our contribution are as follows.

- We propose a universal Few-Shot Learning approach which can be used for significantly different datasets.
- We propose the first One-Shot fake face detection method.
- Our experiments show that, the proposed method is able to detect the majority of test fake faces.

## II. RELATED WORK

In this section, we review the most important researches done in Few-Shot Learning and fake image detection.

### A. Fake Image Detection

Since the invention of GANs, many researches investigate the high-quality image synthesis [16,17,18,19,20]. Recent advances in this field makes the GAN image detection a new research challenge in image forensics. Here, we review limited published papers in this area [3]. Marra et al. [1] investigate the detection of images manipulated by GAN-based image-to-image translation. They found that, detectors perform very well on original images, but some of them show dramatic impairments on Twitter-like compressed images. Valle et al. [2] proposed two different approaches to detect GAN generated images. First, they used statistical analysis and comparison of raw pixel values and extracted features. Second, they prove that, fake samples violate the specifications of the real data. Li et al. [3] analyzed the disparities in color components between real images and fake images. They discussed that, GAN generated images have more obvious differences from real images when they are transformed to other color spaces, such as HSV and YCbCr. McCloskey et al. [4] found that, the frequency of over-exposed pixels provides good discrimination between GAN-generated image and real images. Nataraj et al.

[5] proposed a combination of pixel cooccurrence matrices and deep learning (CMD). Co-occurrence matrices are computed on the color channels of an image and then trained a deep convolutional neural network to distinguish GAN generated fake images from real ones. Hsu et al. [6] proposed a novel deep forgery discriminator (DeepFD) based on embedding the contrasting loss, to detect the fake generated images.

*B. Few-Shot Learning*

Given few training instances of known classes, Few-Shot Learning [8] tries to generalize extracted knowledge to predict the class membership of unseen data. Previous works in this area fall into three categories: Metric Learning, Meta Learning and Data Augmentation [22]. Metric Learning tries to find a metric space suitable for few-shot classification. Koch et al. [7] used Siamese neural networks to find the similarity between inputs. Vinyal et al. [21] introduced Matching Networks for One-Shot Learning via augmenting neural networks with external memories. Prototypical Networks learn a metric space in which classification can be performed by computing distances to prototype representations of each class. Sung et al. [9] proposed relation network to learn an embedding and a deep non-linear distance metric for comparing query and sample items. Relation network is end-to-end with episodic training to find the best distance metric for effective Few-Shot learning. Meta-learning or learning-to-learn gradually learns generic information (meta knowledge) from seen data, and generalizes meta learner for a new task T using task-specific information [22]. Finn et al. [10] proposed an algorithm for meta-learning compatible with any model trained with gradient descent and applicable to a variety of different learning problems, including classification, regression, and reinforcement learning. Hariharan et al.[12] used hallucination and shrinking the features to augment new data. overlaying the samples [13] or randomly erasing parts of images [14] are other practiced approaches to add new instances into available training data. One possible approach to address FSL problem is data augmentation. Augmenting new data to limited training instances is done via flipping, rotating, adding noise and randomly cropping images. Wang et al. [11] mixed a meta-learner with a "hallucinator" to produce additional training example.

III. BACKGROUND

Research community is excited of recent advances in the field of generative models. The examples can be found in different areas including speech synthesis [24] and image to-image translation [25]. However, such strong tools may enable the attackers to challenge the privacy of online models. Since the latent space is low dimensional, a search-based reconstruction attack might be used to find the most similar instance to the target. Among all new powerful generative models, we focus on PG-GAN and TL-GAN.

*A. GAN*

The learning process of the GANs is to train a discriminator $D$ and a generator $G$ simultaneously. The target of $G$ is to learn the distribution $p_g$ over data $x$. $G$ starts from sampling input variables $z$ from a uniform or Gaussian distribution $p_z(z)$, then maps the input variables $z$ to data space $G(z;\theta_g)$ through a differentiable network. $D$ is a classifier $D(x;\theta_d)$ that aims to recognize whether the input is from training data or from $G$. The minimax objective for GANs can be formulated as follows.

$$\min_G \max_D V_{GAN}(D,G) = \mathbb{E}_{x \sim p_x}[logD(x)] + \mathbb{E}_{z \sim q_z}[log(1 - D(G(z)))]$$

The Nash equilibrium of this particular game is achieved at:

$$p_{data}(x) = p_{gen}(x) \quad \forall x$$
$$D(x) = \frac{1}{2} \forall x$$

Training GAN is equivalent to minimizing Jensen-Shannon divergence between generator and data distributions.

*1) Deep Convolutional GANs (DCGANs):* The idea of DCGANs is to replace fully connected hidden layers in standard GANs with convolutions and the pooling layers are replaced with strided convolutions on the discriminator and fractional strided convolutions on the generator. Generally, convolutional nets are useful to find areas of correlation within a vector which is called spatial correlations. All the state of the art face generation models are based on DCGANs.

*2) GAN training:* The training algorithm of GANs is as follows.

---

**Algorithm 1** Minibatch stochastic gradient descent training of generative adversarial nets.

1: **for** number of training iterations **do**
2:   **for** k steps **do**
3:     Sample minibatch of $m$ noise $\{z^{(1)},...,z^{(m)}\}$ sample from noise prior $p_g(z)$.
4:     Sample minibatch of $m$ sample $\{x^{(1)},...,x^{(m)}\}$ from data generating distribution $p_{data}(x)$.
5:     Update the discriminator by ascending its stochastic gradient.
$$\Delta_{\theta_d} \frac{1}{m} \sum_{i=0}^{m} [logD(x^{(i)})] + [log(1 - D(G(z^{(i)})))]$$
6:   **end for**
7:   Sample minibatch of $m$ noise sample $\{z^{(1)},...,z^{(m)}\}$ from noise prior $p_g(z)$.
8:   $\Delta_{\theta_g} \frac{1}{m} \sum_{i=0}^{m} [log(1 - D(G(z^{(i)})))]$
9: **end for**

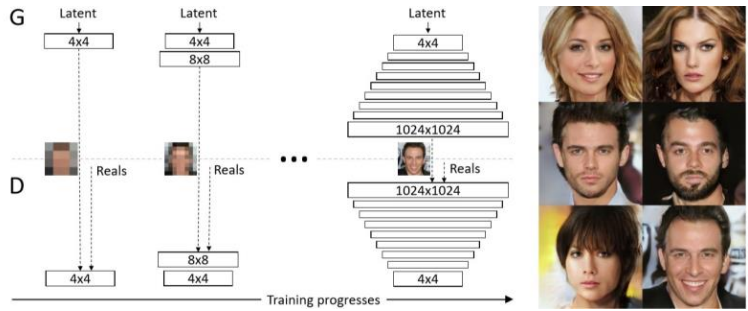

Fig. 1. PG-GAN training architecture: High resolution faces are generated by progressive growing of training data[16].

## B. PG-GAN

The main idea of progressive growing of GAN (PG-GAN) [16] is to start with low-resolution images and then progressively increase the resolution by adding layers to the networks.
The generator and discriminator grow synchronously and layers in both networks remain trainable throughout the training process as shown in Figure 1.

## C. TL-GAN

PG-GAN is so powerful to generate realistic faces. However, it suffers non-transparent latent space. It means that, the user has no idea how to change the noise vector to reach the favorite face. Transparent Latent Space GAN (TL-GAN) [18] was proposed to address this problem. The main idea of TL-GAN is based on this observation that latent space is continuous which lets us to interpolate between two points in the latent space to reach a smooth transition of synthesized faces. The key innovation of TL-GAN is to take advantage of a face classifier based on several face features including gender, age, etc. Once this is done, a link is established between latent vectors and high level face features which makes the latent space transparent as shown in Figure 2. Transparent latent space helps the user to manipulate the face based on face characteristics as shown in Figure 3.

## IV. OUT-OF-CONTEXT OBJECT DETECTION

Scene understanding and object detection methods are not perfect since they may detect a set of meaningful objects in a random image. Yet, they behave consistently meaning that, similar objects are detected in similar images. In this paper, we take advantage of this intuition to extract high level features from the images. Since detected objects may appear out-of-context, we call it out-of-context feature extraction. The core idea of proposed method is to transform an image as numerical entity into a categorical vector via scene understanding methods. Each feature in new space represents the probability of existing an object in the original image. The target vector is sparse since different objects may appear in different instances in training set. Given a set of training instances $(x_1, x_2, \cdots, x_m)$ where $x_i \in R_n$, $m$ sparse vectors are created as $(x'_1, y_1), (x'_2, y_2), \cdots, (x'_m, y_m)$ where $x^0_i \in R^w$ and $w$ is the number words in dictionary.

## A. Fake Face Detection

Unfortunately, the best face recognition systems fail to discriminate fake and real faces. Figure 4, shows two fake faces generated by *thispersondoesnotexist.com*. We analyzed these fake faces by facial analysis tool of Amazon Rekognition [15]. The results show that, they have been recognized as human face with 100 percent confidence as shown in Table 1.

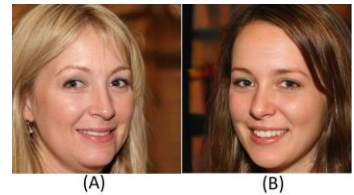

Fig. 4. Two fake faces generated by *thispersondoesnotexist.com*.

TABLE I
CONFIDENCE SCORES RETURNED BY AMAZON REKOGNITION.

|   | Face | Female | Smiling | Age range | No Glasses |
|---|------|--------|---------|-----------|------------|
| A | 100% | 99.7%  | 99.6%   | 23 - 35   | 99.8%      |
| B | 100% | 99.8%  | 99.8%   | 22- 34    | 99.6%      |

Appearance of some objects is natural in face images like clothes, tie and etc. However, detecting some objects are really weird in face images like finger. Table 2 and Table 3 show the object detection results on part (a) and part (b) of Figure 4. The results show that some weird objects are detected in the fake images despite the fact that they are not visible by human eyes. As mentioned earlier, some out-of-context objects are detected in the fake faces including finger which is common between part (a) and part (b) of Figure 4. Given all detected objects in the training set, we can form bag of words. We can also remove the features which are appeared in all instances.

TABLE II
OBJECT DETECTION CONFIDENCE SCORES ON PART (A) OF FIG.4

| Human  | Person | Face   | Decor  | Finger | Tie    |
|--------|--------|--------|--------|--------|--------|
| 99.60% | 99.60% | 99.60% | 63.50% | 61.10% | 55.3 % |

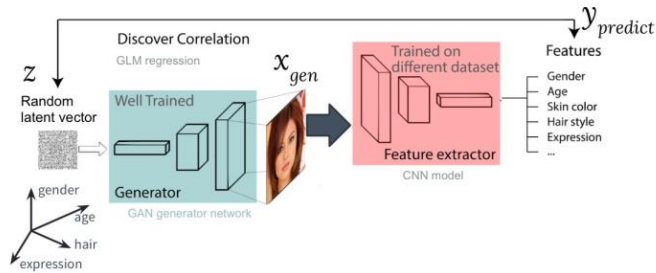

Fig. 2. TL-GAN architecture : A face classifier is used to make the latent space transparent [18].

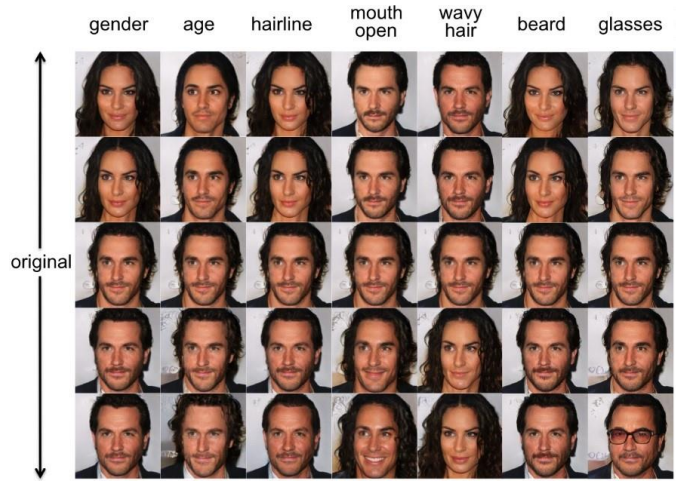

Fig. 3. TL-GAN enables the users to manipulate the generated faces based on face characteristics [18].

TABLE III

OBJECT DETECTION CONFIDENCE SCORES ON PART (B) OF FIG.4

| Human | Person | Face | Clothing | Finger | Wood |
|---|---|---|---|---|---|
| 99.5% | 99.5% | 99.5% | 74.1 % | 59% | 82.4 % |

Our idea to detect GAN generated fake faces is to use object detection methods to analyze the face rather than using face recognition techniques. Our experiments show that, object detection methods can find out-of-context objects in GAN generated fake faces which can help to discriminate the fake faces and real ones given only one training instances.

## V. EXPERIMENTS

In this section, we test the existence of out-of-context objects in the fake faces. Our experiments show that, the out-of-context objects are detected more frequently among low quality faces generated by TL-GAN. Surprisingly, the rate of detected out-of-context objects among fake faces generated by *thispersondoesnotexist.com* is very high despite the fact that all of them are high quality faces. Figure 5 shows some of the collected faces from *thispersondoesnotexist.com*. Table IV shows that finger has been detected in all of them with high confidence scores.

### A. Evaluation

We compared our proposed method with two previous works including CMD [5] and DeepFD [6] to detect the fake faces generated by *thispersondoesnotexist.com*. To do so, we collected 100 fake faces from mentioned website. Our experiments show that, proposed one-shot fake face detection can outperform CMD and DeepFD. Considering the fact that, CMD and DeepFD are using a huge training instance, their efficiency results on high quality fake faces generated by *thispersondoesnotexist.com* is disappointing as shown in Figure 6. Note that, this site takes advantages of Style-GAN [17] to create the fake faces.

## VI. CONCLUSION

In this paper, we proposed a novel One-Shot GAN generated fake face method based on out-of-context object detection. Instead of using face analysis approaches, we used scene understanding methods. We observed that, some out-of-context objects are appeared in fake faces which can help us to discriminate real/fake face images. The proposed method can be used in any other One-Shot anomaly detection areas. As a future work, we are planning to extend the proposed method to address other related problems.

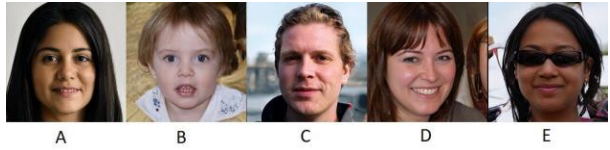

Fig. 5. High quality faces generated by *thispersondoesnotexist.com* : Finger is detected in all these images by object detection model.

TABLE IV
OUT-OF-CONTEXT OBJECT DETECTION PROBABILITIES OF HIGH QUALITY FAKE FACES OF FIGURE 5.

|  | A | B | C | D | E |
| --- | --- | --- | --- | --- | --- |
| Finger | 56.4% | 59.9% | 59.8% | 75.7% | 91.5% |

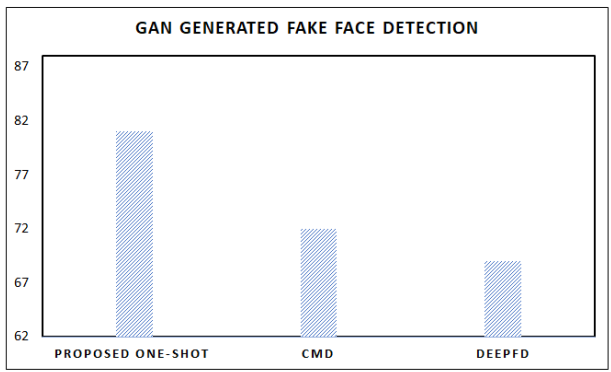

Fig. 6. GAN generated fake face detection accuracy results tested on collected fake faces from *thispersondoesnotexist.com*.

## REFERENCES

[1] Marra, Francesco, et al. "Detection of gan-generated fake images over social networks." 2018 IEEE Conference on Multimedia Information Processing and Retrieval (MIPR). IEEE, 2018.

[2] Valle, Rafael, Wilson Cai, and Anish Doshi. "TequilaGAN: How to easily identify GAN samples." arXiv preprint arXiv:1807.04919 (2018).

[3] Li, Haodong, et al. "Detection of deep network generated images using disparities in color components." arXiv preprint arXiv:1808.07276 (2018).

[4] McCloskey, Scott, and Michael Albright. "Detecting gan-generated imagery using color cues." arXiv preprint arXiv:1812.08247 (2018).

[5] Nataraj, Lakshmanan, et al. "Detecting GAN generated fake images using co-occurrence matrices." arXiv preprint arXiv:1903.06836 (2019).

[6] Hsu, Chih-Chung, Chia-Yen Lee, and Yi-Xiu Zhuang. "Learning to Detect Fake Face Images in the Wild." 2018 International Symposium on Computer, Consumer and Control (IS3C). IEEE, 2018.

[7] Koch, Gregory, Richard Zemel, and Ruslan Salakhutdinov. "Siamese neural networks for one-shot image recognition." ICML deep learning workshop. Vol. 2. 2015.

[8] Snell, Jake, Kevin Swersky, and Richard Zemel. "Prototypical networks for few-shot learning." Advances in neural information processing systems. 2017.

[9] Sung, Flood, et al. "Learning to compare: Relation network for few-shot learning." Proceedings of the IEEE Conference on Computer Vision and Pattern Recognition. 2018.

[10] Finn, Chelsea, Pieter Abbeel, and Sergey Levine. "Model-agnostic metalearning for fast adaptation of deep networks." Proceedings of the 34th International Conference on Machine Learning-Volume 70. JMLR. org, 2017.

[11] Wang, Yu-Xiong, et al. "Low-shot learning from imaginary data." Proceedings of the IEEE conference on computer vision and pattern recognition. 2018.

[12] Hariharan, Bharath, and Ross Girshick. "Low-shot visual recognition by shrinking and hallucinating features." Proceedings of the IEEE International Conference on Computer Vision. 2017.

[13] Inoue, Hiroshi. "Data augmentation by pairing samples for images classification." arXiv preprint arXiv:1801.02929 (2018).

[14] Zhong, Zhun, et al. "Random erasing data augmentation." arXiv preprint arXiv:1708.04896 (2017).

[15] https://aws.amazon.com/rekognition/

[16] Karras, Tero, et al. "Progressive growing of gans for improved quality, stability, and variation." arXiv preprint arXiv:1710.10196 (2017).

[17] Karras, Tero, Samuli Laine, and Timo Aila. "A style-based generator architecture for generative adversarial networks." Proceedings of the IEEE Conference on Computer Vision and Pattern Recognition. 2019.

[18] https://github.com/SummitKwan/transparent latent gan.

[19] Mansourifar, Hadi, Lin Chen, and Weidong Shi. "Virtual Big Data for GAN Based Data Augmentation." 2019 IEEE International Conference on Big Data (Big Data). IEEE, 2019.

[20] Goodfellow, Ian, et al. "Generative adversarial nets." Advances in neural information processing systems. 2014.

[21] Vinyals, Oriol, et al. "Matching networks for one shot learning." Advances in neural information processing systems. 2016.

[22] Wang, Yaqing, and Quanming Yao. "Few-shot learning: A survey." arXiv preprint arXiv:1904.05046 (2019).

[23] Mansourifar, Hadi and Weidong Shi. "Deep Synthetic Minority OverSampling Technique." arXiv:2003.09788 (2020).

[24] Kaneko, Takuhiro, et al. "Generative adversarial network-based postfilter for statistical parametric speech synthesis." 2017 IEEE International Conference on Acoustics, Speech and Signal Processing (ICASSP). IEEE, 2017.

[25] Yi, Zili, et al. "Dualgan: Unsupervised dual learning for image-toimage translation." Proceedings of the IEEE international conference on computer vision. 2017.